\documentclass{llncs}
\usepackage{times}
\usepackage{soul}
\usepackage{url}
\usepackage[hidelinks]{hyperref}
\usepackage[utf8]{inputenc}
\usepackage[small]{caption}
\usepackage{amsmath}
\usepackage{amsfonts}
\usepackage{algorithm}
\usepackage{graphicx}
\usepackage{float}
\usepackage{bm}
\usepackage{amsmath}
\usepackage{tikz}

\usepackage[noend]{algpseudocode}

\title{Improving generalization in reinforcement learning through forked agents} 
\author{
Olivier Moulin\inst{1}
\and
Vincent Francois-Lavet\inst{1}\and
Paul Elbers\inst{2}\and
Mark Hoogendoorn\inst{1}}
\institute{Vrije Universiteit Amsterdam, Department of Computer Science\and Vrije Universiteit Medical Center Amsterdam, Department of Intensive Care}

\begin{document}
\maketitle
\begin{abstract}

An eco-system of agents each having their own policy with some, but limited, generalizability has proven to be a reliable approach to increase generalization across procedurally generated environments. In such an approach, new agents are regularly added to the eco-system when encountering a new environment that is outside of the scope of the eco-system. The speed of adaptation and general effectiveness of the eco-system approach highly depends on the initialization of new agents. In this paper we propose different initialization techniques, inspired from Deep Neural Network initialization and transfer learning, and study their impact. 
\end{abstract}

\section{Introduction}
\label{Introduction}
Generalization of reinforcement learning (RL) agents to previously unseen environments is a key topic in reinforcement learning. Reinforcement learning agents have the tendency to overfit the environment on which they are trained. 
This problem has been highlighted often in the literature, for example by Cobbe et al. \cite{KCobbe2019} and Packer et al. \cite{CPacker2018}.

The eco-system approach described by Moulin et al. \cite{OMoulin2021} is one of the approaches put forward to improve generalization across environments while maintaining performance on previously seen environments. It is based on an eco-system of agents with the idea that each agent has its own policy with some generalizability, where the combination makes up a highly generalizable system. When a new environment is encountered, existing agents are used, or a new agent is trained when none performs satisfactorily. While the approach is unique in its ability to avoid  catastrophic forgetting, it requires a lot of access to the environment to achieve its goal. 

In this paper, we aim to improve this approach. Hereby, we focus on initialization procedures for new agents. Better initialization has the potential to reduce the heavy burden of ample access to the environments and can additionally improve generalizability. Drawing inspiration from papers on initialization techniques for Deep Neural networks (e.g. Boulila et al. \cite{WBoulila2021}) and on transfer learning (e.g. Taylor et al. \cite{MTaylor2009}, we consider the following initialization options: (i) initialization with the best agent in the pool on the new environment, (ii) with a random agent chosen from a pool, (iii) with an agent not included in the pool and trained on all past environments we refer to as forked agent (iv) and with no initialization at all, which matches the setup from Moulin et al. \cite{OMoulin2021}. 
We evaluate the performance of our innovations in the well-known minigrid environment~\cite{gym_minigrid} and compare to existing state-of-the-art methods.

Our contributions are:
\begin{itemize}
\item Identifying the impact of different initialization techniques on the speed of learning and usage of resources for newly encountered environments.
\item Proposing a new setup improving generalization in reinforcement learning. 
\end{itemize}
This paper is organized as follows. Section \ref{related_work} presents the related work which has inspired the approaches tested in this paper. Section \ref{approach} provides an explanation of our approach. Next, section \ref{experimental_setup} presents the experimental setup used to evaluate the approach. Section \ref{results} presents the results of the experiments. We end with a discussion in  section \ref{discussion_conclusion}.

\section{Related work}
\label{related_work}
\subsection{Eco-system approach}
The concept of an eco-system to improve generalizability  was recently introduced by Moulin et al. \cite{OMoulin2021} where they proposed to use a set of specialist agents (trained on only one environment) to achieve better generalization at the group level (eco-system).

\subsection{Generalization in Reinforcement Learning}
The way to assess generalization of a given RL system and the usage of procedurally generated environments is inspired by the papers from Cobbe et al. \cite{KCobbe2020} and \cite {KCobbe2019}.
Several papers are focused on improving generalization in the Reinforcement Learning context.
In addition to the newly proposed eco-system approach referenced above, two other main categories can be found.
The first category is focused on creating a representation of the environment. This approach helps learning a good policy and reducing at the same time over-fitting. (e.g. Sonar et al. \cite{asonar2020})
The second category is focused on adding noise and/or information bottlenecks in the Neural Network.
This approach also reduce the risk of over-fitting to the training environment. (e.g. Chen \cite{jzchen2020}, Lu et al. \cite{Xlu2020} and Igl et al. \cite{Migl2019})
\subsection{Specialist and generalist agents to improve training}
Zhiwei et al. \cite{JZhiwei2022} proposes to use a set of specialist agents with imitation learning techniques to improve the learning capabilities of a generalist agent when additional training steps on it does not help improving the accuracy.
After leveraging the specialist agents to train the main generalist agent by imitation learning, it resumes the training of the generalist agent by normal accesses to the environment.
This approach offers some similarities to our approach by using very specialized agents as well as leveraging the generalizability of one agent being trained on multiple environments, but it also differs from our forked agent approach where we use one agent trained on multiple environment (generalist agent) to initialize the agents of the eco-system (specialist agents).
\subsection{Initialization of Neural Network / agents}
In this paper we show that the initialization of the agents (Neural Network / Policy of the agent) in an eco-system setup has an impact on the performance of the overall system.
The ideas for the different initialization techniques presented in this paper have been inspired by a survey paper from Boulila et al. \cite{WBoulila2021} where they list the most used techniques to initialize the weights of a neural network as well as another survey paper done in the area of transfer learning from Taylor et al. \cite{MTaylor2009}, which gives a good overview of the domain. These papers have been used as inspiration and have helped us trigger new ideas on how to improve the initialization of the new agents. These papers are also quite different from our approaches in the fact that they don't relate specifically to Reinforcement Learning or the eco-system setup.
\section{Approach}
\label{approach}
The eco-system approach, described by Moulin et al. \cite{OMoulin2021}, is based on the assumption that each agent trained on an environment is able to generalize a bit , which allows it to perform properly on a limited number of other environments. 
The eco-system is composed of  multiple agents (a pool of agents).
Each agent part of the eco-system is trained as a standard RL agent, but only on one environment (specialist).
The generalization improvement is made by gathering the generalization capabilities of each agent.
In this paper we look at different techniques to improve the overall performance of this approach, thereby focusing on the initialization of new agents.
The performance increase is defined as any action which leads to improving how the overall system generalizes to new environments (better generalization index) as well as how it reduces the resources needed (number of agents in memory, number of training cycles needed) to accomplish the same or better level of generalization.

\subsection{Reinforcement Learning formulation}
Reinforcement Learning is based on the interactions between an agent and an environment on which is it trained.
These interactions happen over discrete time-steps.
The data used to train the agent is gathered directly from the environment at the same time (like in our case), or at a later time (like when using replay memory) it is explored by the agent, which makes it a different approach from other Machine Learning techniques, like supervised learning, for example, where all the training data is provided upfront to the agent before it starts interacting with the environment.
The environment is formalized as an MDP (Markov Decision Process) defined by (i) a state space (composed of all the potential observations from the environment which can be gathered by the agent), called $\mathcal S$ which can be continuous or not, (ii) an action space (composed of all the potential actions which can be taken by the agent), called $\mathcal A=\{1, \ldots, N_{\mathcal A}\}$, (iii) a transition function (allowing to move from one state to a new one according the action selected) noted $T:~\mathcal S \times \mathcal A  \to \mathbb P(\mathcal S)$, and (iv) the reward function (providing the reward gathered when selecting a given action from a given state) , noted $R:~\mathcal S \times \mathcal A \times \mathcal S \to \mathcal R$ where $\mathcal R$ encompass all the possible rewards in a range $R_{\text{max}} \in \mathbb{R}^{+}$ ($[0,R_{\text{max}}]$).
After initialization of the MDP, noted $M$ the agent has access to a distribution of initial states, noted $b_0(s)$.
At each time step $t$, the agent will select an action available in the current state of the system, noted $s_t \in \mathcal S$. This action is part of the policy $\pi:\mathcal S \times \mathcal A \rightarrow [0,1]$.
Taking the selected action $a_t \sim \pi(s_t,\cdot)$, will move the agent in a new state which it can observe, noted $s_{t+1} \in \mathcal S$, and will grant the agent a given reward signal noted $r_t \in \mathcal R$.
\subsection{Proximal Policy Optimization formulation}
Different algorithms can be used to implement the Reinforcement Learning approach : DDQN (\cite[Van Hasselt et al.]{HVanHasselt2019}), Actor-Critic (\cite[Konda et al., 1999]{Vkonda1999}), PPO (\cite [Schulman et al., 2017] {jschulman2017}), etc. In this paper we choose to focus on the Proximal Policy Optimization (PPO) to match what was done by Moulin et al. \cite{OMoulin2021}, as our goal is to show how optimizing the initialization of the agents can increase the overall performance of the approach. 
The Proximal Policy Optimization (PPO) algorithm (cf. \cite [Schulman et al., 2017] {jschulman2017}) is an improvement of the actor-critic method (\cite[Konda et al., 1999]{Vkonda1999}).
The parameters, noted $w$ of a given policy $\pi_{w} (s,a)$ are updated to optimize $A^{\pi_{w}}(s,a)=Q^{\pi_{w}}(s,a)-V^{\pi_{w}}(s)$. 
The PPO algorithm adds a limit on the policy changes to reduce instability and avoid too much variation after each training step.
This results in maximizing the following objective in expectation over $s~\sim~\rho^{\pi_w}, a~\sim~\pi_w$.
Where :
$$\min \Big( r_t(w) A^{\pi_w}(s,a), \text{clip} \big(  r_t(w), 1-\epsilon, 1 + \epsilon \big) A^{\pi_w}(s,a)  \Big)
$$
\begin{itemize}
    \item $r_t(w)=\frac{\pi_{w+\bigtriangleup w} (s,a)}{ \pi_{w} (s,a)}$,
    \item $\rho^{\pi_{w}}$ being the discounted state distribution.
    \item being defined as
\\$\rho^{\pi_{w}}(s)=\sum_{t=0}^{\infty} \gamma^t Pr\{s_t=s | s_0, {\pi_{w}} \}$
    \item $\epsilon \in \mathbb R$ being a hyper-parameter.
\end{itemize}
The implementation of the PPO algorithm we are using in this paper is stable-baselines3 (cf. \cite [OpenAI]{sb3PPO}).

\subsection{Eco-system algorithm}
The ecosystem algorithm, described by  Moulin et al. \cite{OMoulin2021} works as follows:
Each time the eco-system meets a never seen environment, it will browse its pool of agents and try to find one that can solve the environment (solving means harvesting a reward greater than a given threshold). If no agent from the pool can be found, then a new agent will be created, trained on the new environment and added to the pool. The eco-system will then check if this new agent can replace an existing agent in the pool, and if it is the case remove the old agent from the pool.
The generalization of the eco-system is accomplished by all the agents in the pool.
We summarized it in algorithm \ref{codetrainingmultiagent} as a baseline to the modifications detailed in the next sections. 
The non colored part of the algorithm correspond to the original eco-system algorithm. The colored parts are used to highlight the changes made for each initialization technique.
\begin{algorithm}[!htbp]
	\caption{eco-system - learn($M_{i})  // updated\ from\ Moulin\ et\ al.\ original$} 
	\label{codetrainingmultiagent}
    \begin{algorithmic}
    \State $e^{*}$ $\leftarrow$ $\emptyset$ \textit{\scriptsize \ \#good enough agent found}
	\State $n$ $\leftarrow$ 0 \textit{\scriptsize\ \#loop var.}
	\While{$e^{*}$ = $\emptyset$ and $\bigcup$ $e_{0...n}$ $\neq$ $\mathcal{E}$} 
        \State \textit{\scriptsize\ \#while good policy not found }
		\State \textit{\scriptsize\ \#and not all agents reviewed}
		\State $\mathcal{R}^{\pi_{e_{n}}}_{M_{i}}=$ \text{test\_agent}($e_{n}$,$M_{i}$)
        \State \textit{\scriptsize \ \#Total reward from $e_{n}$ on $M_{i}$}
		\If {$\mathcal{R}^{\pi_{e_{n}}}_{M_{i}}$ $\geq$ $l$ } \textit{\scriptsize \#if $e_{n}$ solve $M_{i}$}
			\State $e^{*}$ $\leftarrow$ $e_{n}$ \textit{\scriptsize \ \#good enough agent found = $e_{n}$}
		\Else 
            \State $n$ $\leftarrow$ $n+1$
        \EndIf
	\EndWhile
    \color{orange}
    \State The following while statement replace the previous one
    \State $best\_performing$ $\leftarrow$ 0 \textit{\scriptsize\ \#loop var.}
    \State $best\_reward$ $\leftarrow$ 0 \textit{\scriptsize\ \#loop var.}
    \While{$\bigcup$ $e_{0...n}$ $\neq$ $\mathcal{E}$} 
        \State \textit{\scriptsize\ \#while good policy not found }
		\State \textit{\scriptsize\ \#and not all agents reviewed}
		\State $\mathcal{R}^{\pi_{e_{n}}}_{M_{i}}=$ \text{test\_agent}($e_{n}$,$M_{i}$)
        \State \textit{\scriptsize \ \#Total reward from $e_{n}$ on $M_{i}$}
		\If {$\mathcal{R}^{\pi_{e_{n}}}_{M_{i}}$ $\geq$ $l$ } \textit{\scriptsize \#if $e_{n}$ solve $M_{i}$}
			\State $e^{*}$ $\leftarrow$ $e_{n}$ \textit{\scriptsize \ \#good enough agent found = $e_{n}$}
		\Else 
            \State $n$ $\leftarrow$ $n+1$
        \EndIf
        \If {$\mathcal{R}^{\pi_{e_{n}}}_{M_{i}}$ $\geq$ $best\_reward$ } \textit{\scriptsize \#if $e_{n}$ beats previous best agent}
           \State $best\_performing$ $\leftarrow$ n
           \State $best\_reward$ $\leftarrow$ $\mathcal{R}^{\pi_{e_{n}}}_{M_{i}}$
         \EndIf  
	\EndWhile
    \color{black}
	\If {$e^{*}$ = $\emptyset$} \textit{\scriptsize \#if $e^{*}$ not found}
		\State	$e$ $\leftarrow$ new\_agent()
        \color{purple}
        \State $e$.Neural\_Network $\leftarrow$ $e_{randomly\_chosen}$.Neural\_Network
        \color{orange}
        \State $e$.Neural\_Network $\leftarrow$ $e_{best\_performing}$.Neural\_Network
        \color{green}
        \State $e$.Neural\_Network $\leftarrow$ $Main\_Agent$.Neural\_Network
        \color{black}
		\While{$\mathcal{R}^{\pi_e}_{M_{i}}$ $\leq$ $l$} \textit{\scriptsize \ \#while $e$ cannot solve $M_{i}$}
		    \State learn-epoch($e$,$M_{i}$) 
            \State $\mathcal{R}^{\pi_e}_{M_{i}}=$ \text{test\_agent}($e$,$M_{i}$)
	    \EndWhile
		\State $\mathcal{E}$ $\leftarrow$ $\mathcal{E}$ + $e$ \textit{\scriptsize \ \#add $e$ to the pool}
        \color{green}
        \State $Main\_Agent$.Neural\_Network $\leftarrow$ $e$.Neural\_Network
        \color{black}
		\State \textit{\scriptsize \ \#Following For statement is optional}
		\State \textit{\scriptsize \ \#Only needed if optimization of the pool is needed}
		\For {$f$ $\in$ $\mathcal{E}$} \textit{\scriptsize \#for all agent $f$ in the pool}
			\For {$w$ $\in$ $\delta^{f}$} \textit{\scriptsize \#for all env. $w$ solved by $f$}
				\State $\mathcal{R}^{\pi_{e^{*}}}_{w}=$ \text{test\_agent}($e^{*}$,$w$)
                \If {$\mathcal{R}^{\pi_{e^{*}}}_{w}$ $\geq$ $l$} \textit{\scriptsize \#if $e$ can solve $w$}
				    \State $\delta^{e}$ $\leftarrow$ $\delta^{e}$+$w$ \textit{\scriptsize \#add $w$ to $e$ list}
				\EndIf
			\EndFor
			\If {$\delta^{f}$ $\in$ $\delta^{e}$ }
				\State \textit {\scriptsize \#if $e$ can solve all env. of $f$}
				\State $\mathcal{E}$ $\leftarrow$ $\mathcal{E}$ - $f$ \textit{\scriptsize \#remove $f$ from pool}
			\EndIf
		\EndFor
		\State Sort $\mathcal{E}$ by size $\delta$ descending order
	\Else
		\State $\delta^{e^{*}}$ $\leftarrow$ $\delta^{e^{*}}$+$M_{i}$ \textit{\scriptsize \ \#add $M_{i}$ to $e^{*}$ list}
	\EndIf		
    \end{algorithmic}
\end{algorithm}

\subsection{Initialization techniques}
In order to improve the performance of the eco-system, we have focused our work on finding  a better way to initialize the agents when they are created, trained and added to the pool.
\begin{figure}[!htbp]
	\centering
	\includegraphics[width=5cm]{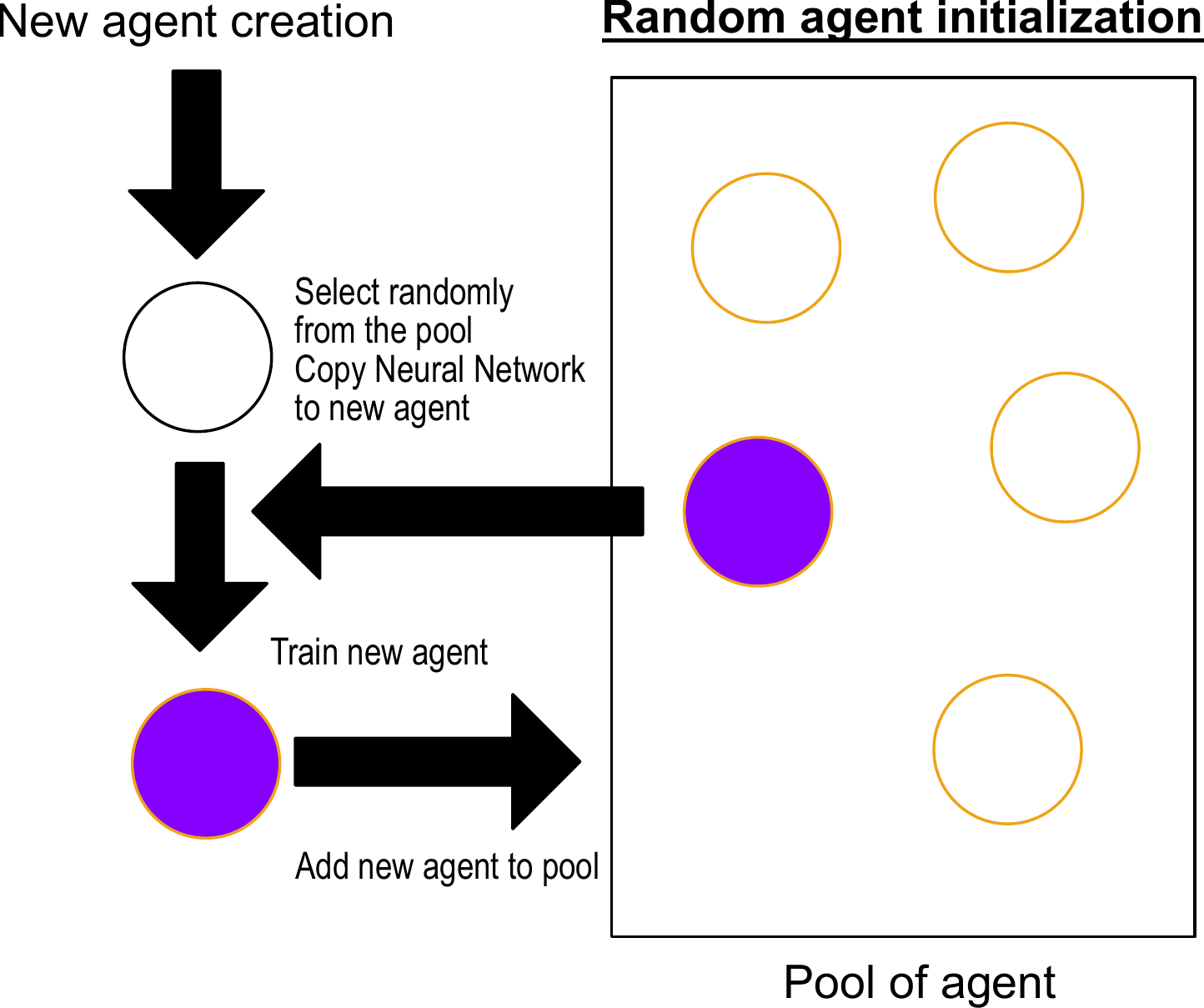}
	\includegraphics[width=5cm]{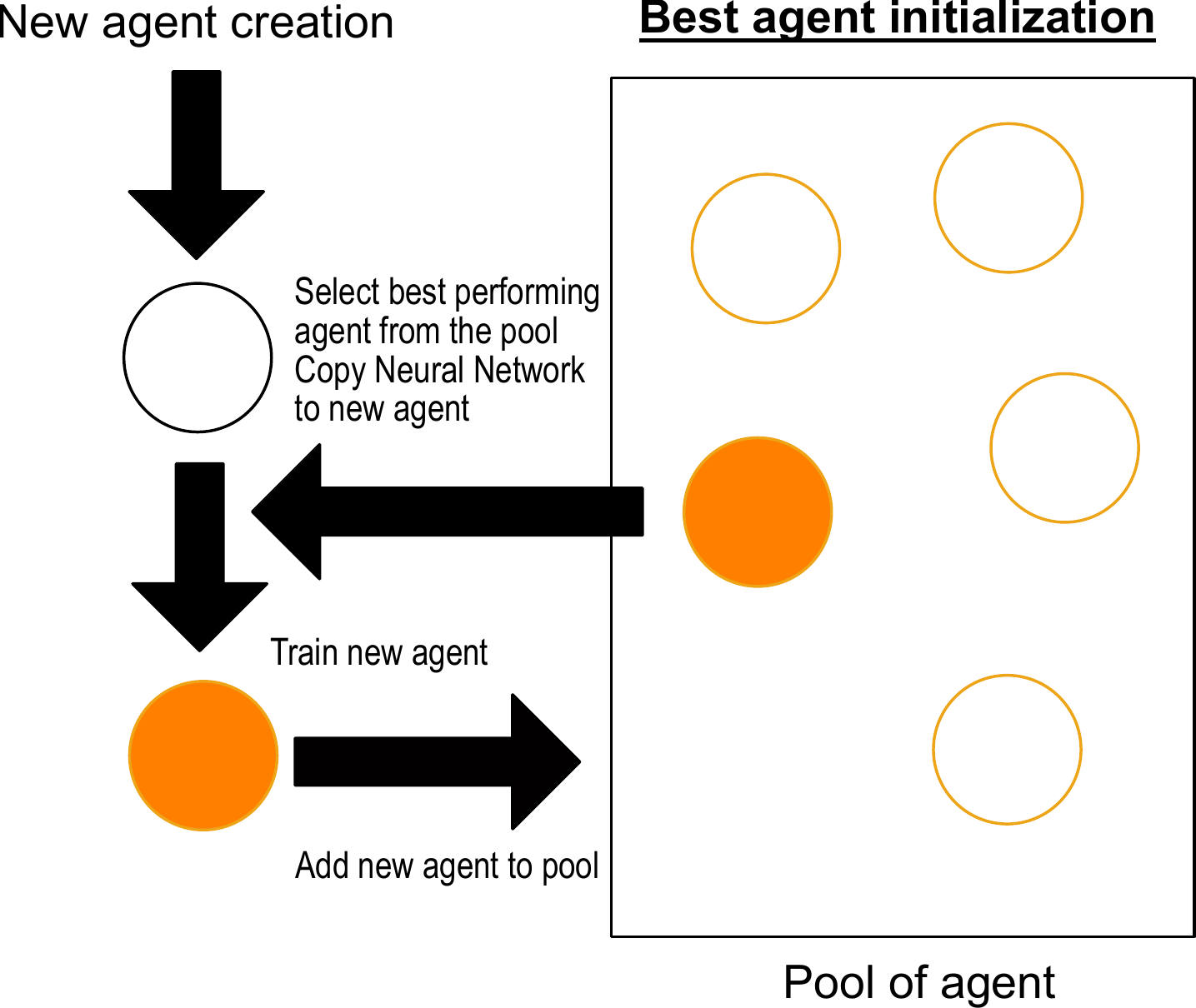}
 	\includegraphics[width=5cm]{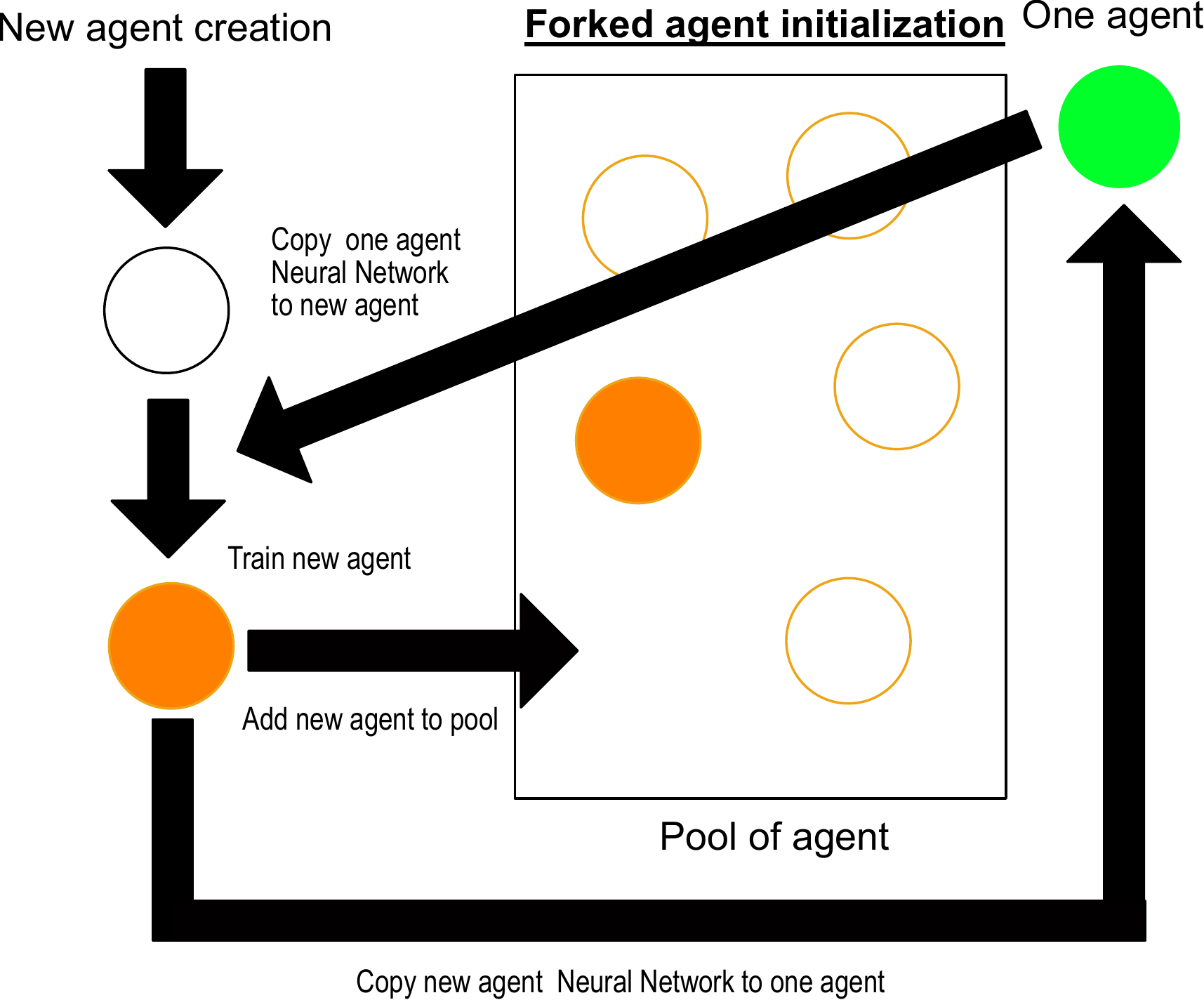}
	\caption{Random, Best and One agent initialization techniques}
	\label{agent_init}
\end{figure}
\paragraph{Basic initialization}
This is the initialization introduced with the eco-system approach. The agent is simply created before being trained on the new environment and then added to the pool of agents.
\paragraph{Random initialization}
With this approach (Figure \ref{agent_init}), each time an agent is created, its neural network is copied from another agent randomly chosen from the pool of agents. The new agent initialized this way is then trained on the new environment and added to the pool of agents. 
The changes in the code are highlighted in algorithm \ref{codetrainingmultiagent} in \color{purple}purple\color{black}.

\paragraph{Best agent initialization}
In the eco-system a new agent is created only if no agent from the pool was able to solve the new environment (reaching the threshold). With this approach (Figure \ref{agent_init}), when testing if an existing agent can solve the new environment, the agent tested which performed the best (while still performing below the desired standard) is stored. When initializing the new agent, the neural network from the best performing agent is used. In case we have multiple agents candidate for the best agent (identical reward), the first one encountered while browsing the pool of agent is selected. The new agent initialized this way is then trained on the new environment and added to the pool of agents.
For this initialization approach, we modify the initial algorithm (algorithm \ref{codetrainingmultiagent}) in two places. 
The changes in the code are highlighted in algorithm \ref{codetrainingmultiagent} in \color{orange}orange\color{black}.
\paragraph{Forked agent initialization}
With this approach (Figure \ref{agent_init}), we create a new agent, called Main Agent outside of the pool of agents.
\begin{figure}[!htbp]
	\centering
	\includegraphics[width=4cm]{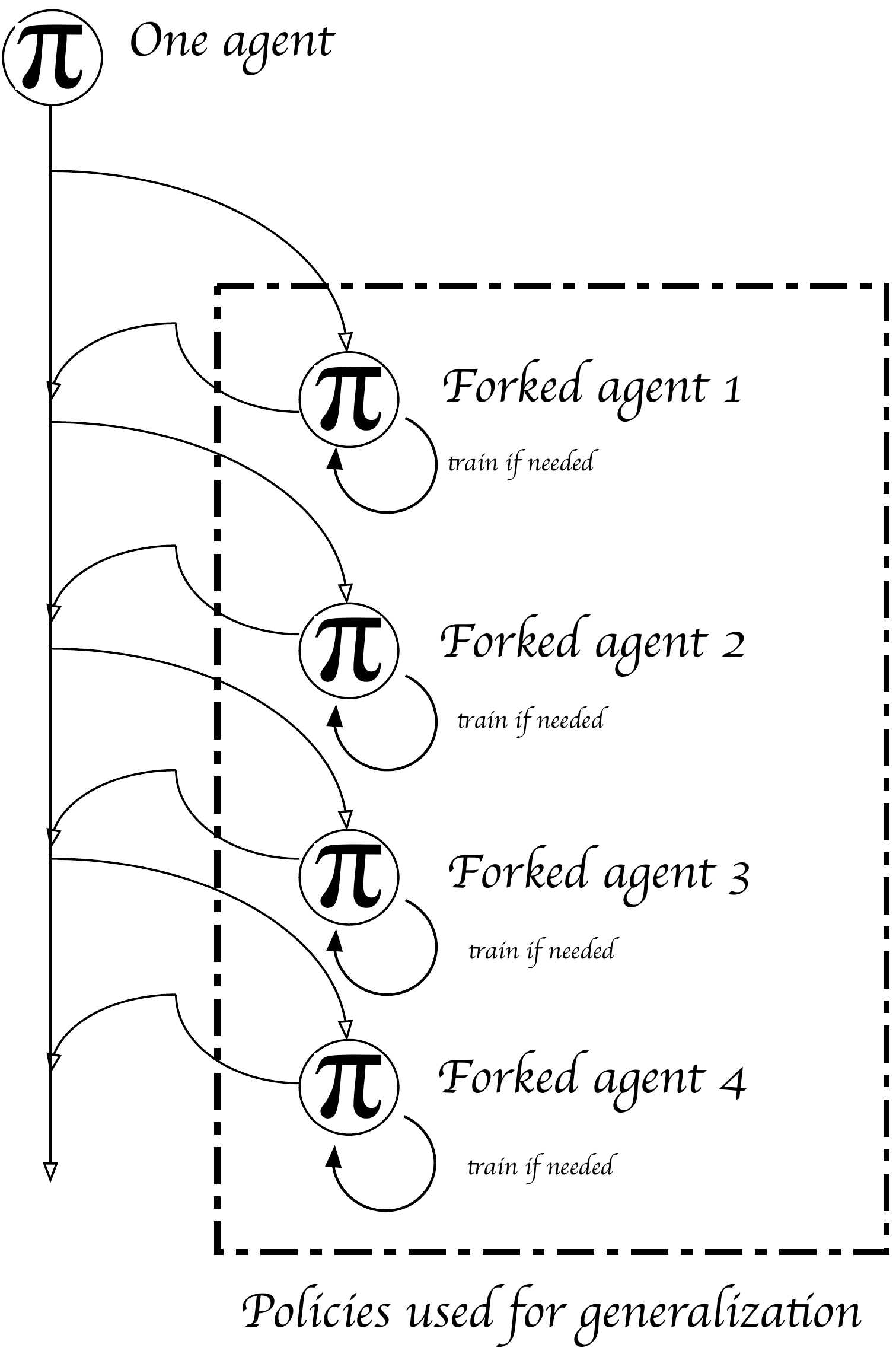}
	\caption{Forked Agent setup}
	\label{forked_policies_concept}
\end{figure}
The Neural Network of the Main Agent is used to initialize each new agent created before to be added to the pool of agents. This is done by creating a fork (or copy) of the Main Agent Neural Network weights and assign them to the weights of the Neural Network of the new agent.  (Figure \ref{forked_policies_concept})
The new agent is then trained on the specific environment for which it has been created, and added to the pool of agents.
The Neural Network of the newly created agent then replace the Neural Network of the Main Agent.
The agent outside of the pool of agents is then trained on all the environments where additional training / agents in the pool are needed.
The Main Agent outside of the pool of agents used for initialization of new agents can forget previously learned environment, but it is not an issue as it stays outside of the eco-system and is not used for generalization purpose.
The changes in the code are highlighted in algorithm \ref{codetrainingmultiagent} in \color{green}green\color{black}.

\section{Experimental setup}
\label{experimental_setup}
The experimental setup we use in this paper is similar to the one used in the paper from Moulin et al. \cite{OMoulin2021}, in order we can assess the performance enhancement of our initialization techniques and our new proposed setup.
\subsection{Environments}
The experiments are conducted on Minigrid.
Minigrid is a commonly used environment in Reinforcement Learning to test algorithms.
We use the FourRooms setup from Minigrid (cf. \cite[Chevalier-Boisvert et al., 2018] {gym_minigrid}). 
Minigrid FourRooms is a  procedurally generated environment, which means in our case that the map, start position, goal position, and obstacles are positioned randomly according to the seed of each level, while the other components of the experiments like reward given to the agents are kept the same for each level.
\begin{figure}[!htbp]
	\centering
	\includegraphics[width=4cm]{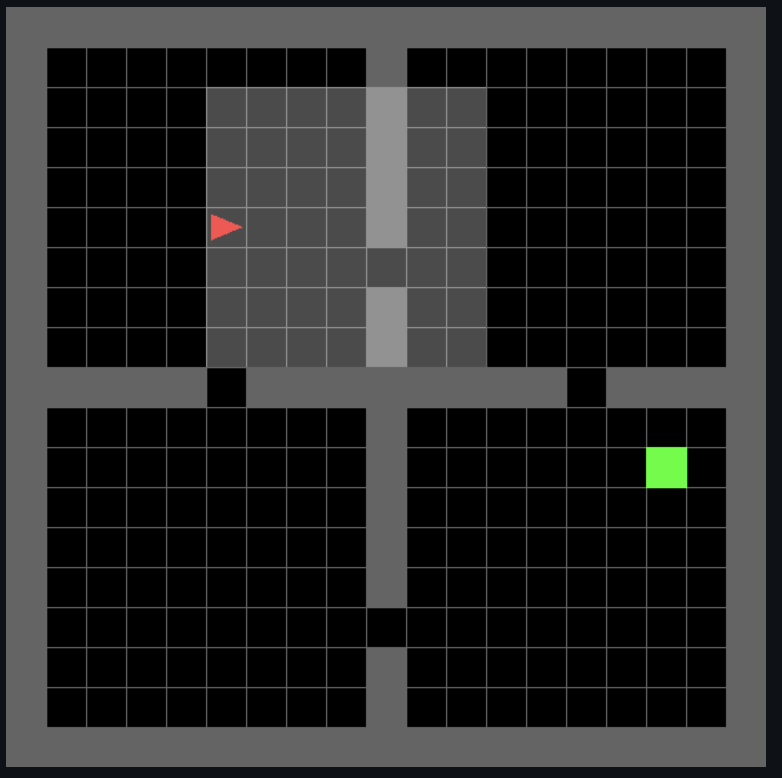}
	\caption{Minigrid FourRoom and Multiroom environment}
	\label{minigrid}
\end{figure}
In these environments we use the basic view of Minigrid which returns as state a partially observable view of the environment using a compact encoding, composed of 3 input values per visible grid cell.
The total structure returned is a 3D table with 7x7x3 values. These values are not pixels.
The 7x7 area represent the part of the environment visible from the agent. The 3 values are a code representing the configuration of each cell.
The agent gets a reward for reaching the goal in the maze. The reward is defined as :
$$1 - 0.9 * (stepsUsed / maxStepsAllowed)$$
\subsection{Performance metrics}
In order to asses the performance of each initialization technique, we have used the adaptability index based on the average reward metric. This metric is based on testing on a number of unseen environments $M_{i} \in \mathcal{M}$. It express the average reward $\mathcal{R}$ gathered by the approach on these environments.
The adaptability index (introduced by Moulin et al. \cite{OMoulin2021}) based on the average reward gathered, noted as $\zeta$, is indicated as a float value, showing the average of total rewards $\mathcal{R}$ gathered over all the new environments $M_{i} \in \mathcal{M}$ on which the approach was tested. It is formalized as follows:
\begin{center}
	$\zeta = \frac{\sum_{i=0}^{n} R_{M_{i}}}{n}$
\end{center}
The metrics are calculated periodically after each approach has been presented to 50 additional environments (and completed the associated training if necessary).
The number of training steps necessary to solve an initial set of 500 environments has been periodically gathered to assess if one approach allows to reduce the computational requirements.
The last performance metrics is the number of agents needed to solve the initial set of 500 environments.
The hyper-parameters are the ones defined by default for stable-baselines3.
The threshold used to indicate that an environment has been solved is a reward of 0.8.

\subsection{Experiments result gathering}
For all experiments, the results are gathered after running 5 experiments with each proposed technique. 

\section{Results}
\label{results}
\begin{figure}[!htbp]
	\centering
	\includegraphics[width=5cm]{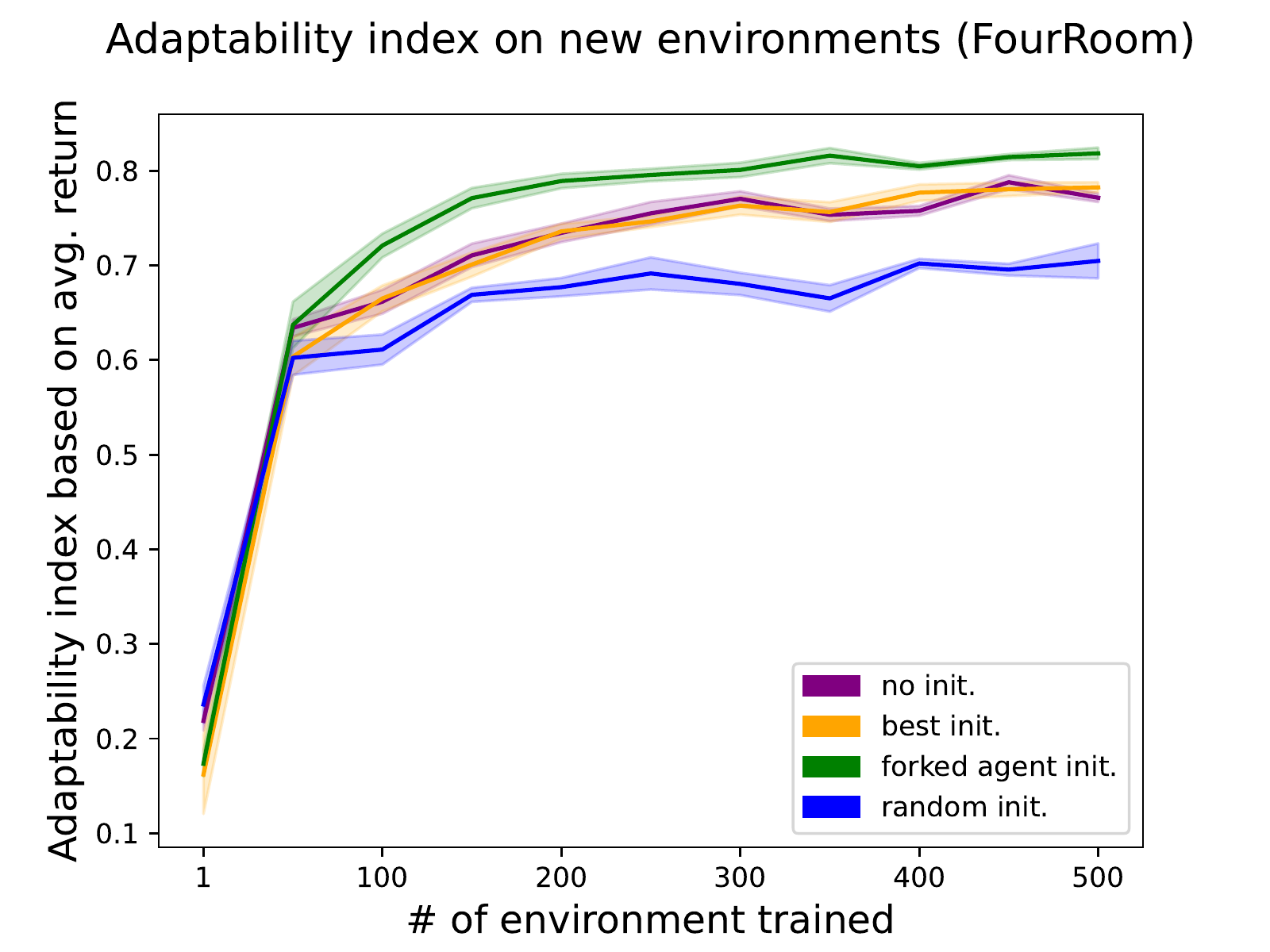}
    \includegraphics[width=5cm]{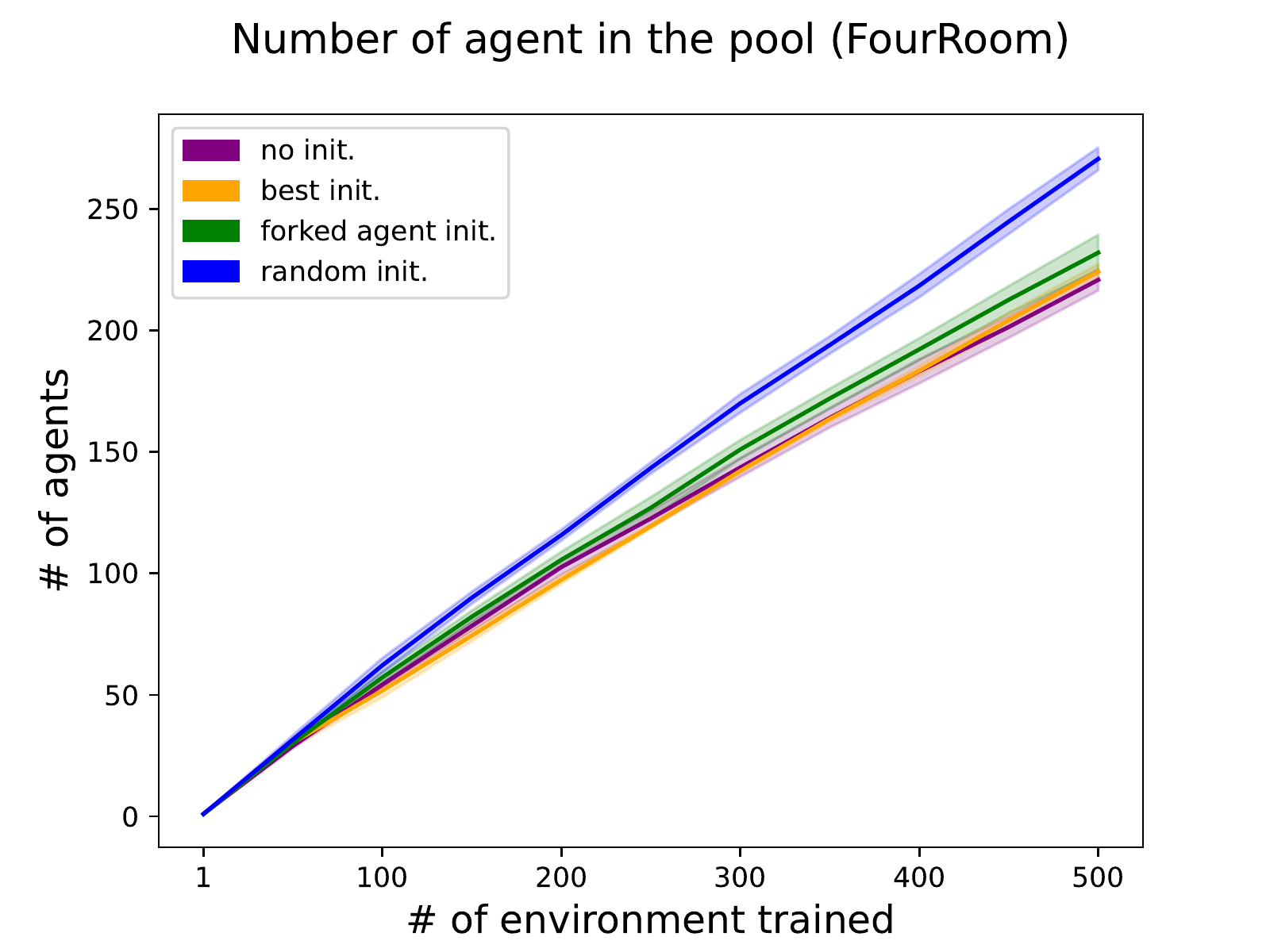}
 	\includegraphics[width=5cm]{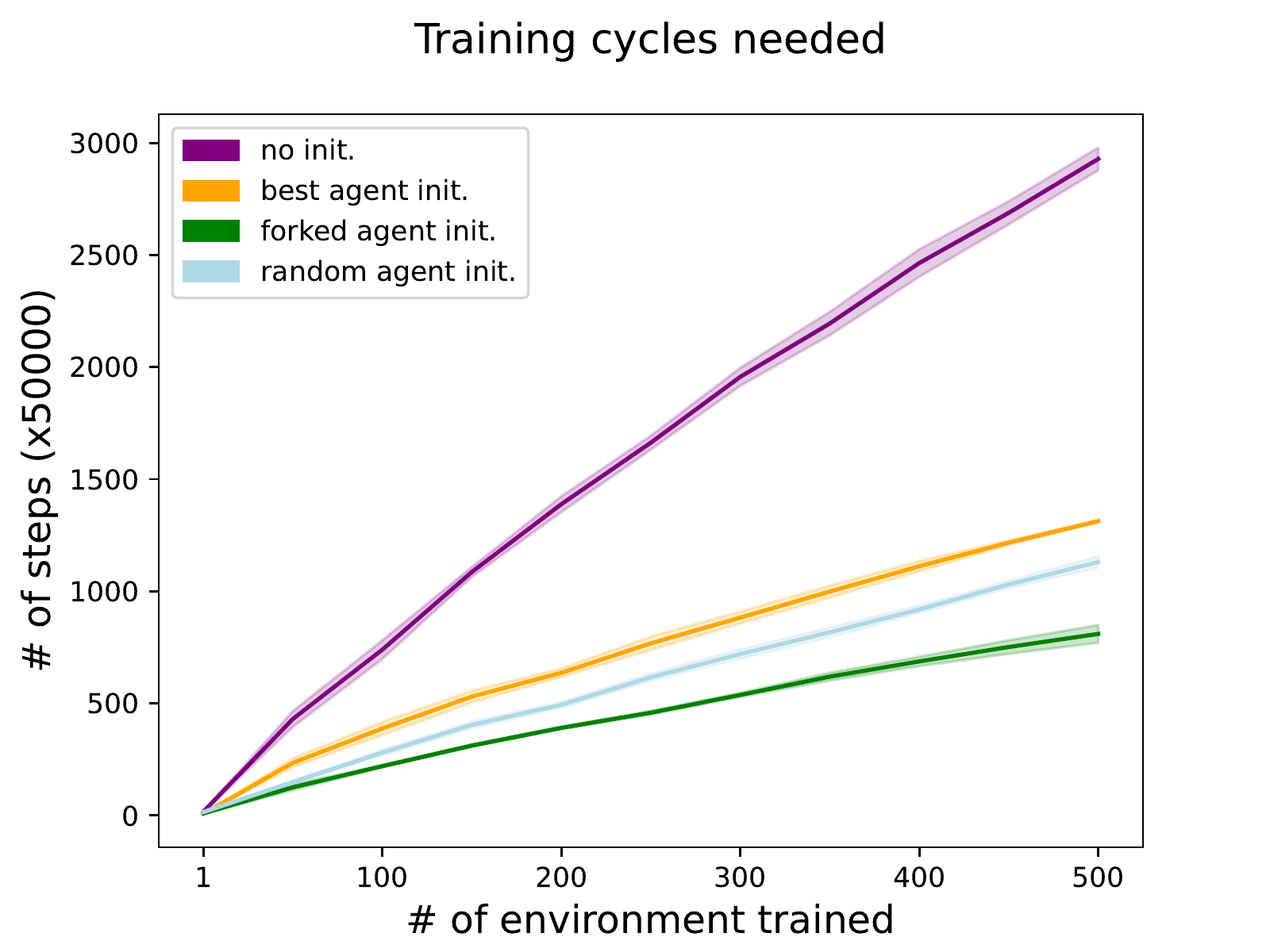}
	\caption{Initialization techniques results and standard error (shaded area)}
	\label{init_results}
\end{figure}

Below we discuss the results we obtained in our experiments.
The charts displayed in this section are showing the average of each indicator as well as the standard error based on the 5 runs.
\paragraph{Adaptability index based on average reward}
We can see in Figure \ref{init_results} that the forked agent initialization technique performs better than all other initialization approaches. 
This performance increase is shown by the average reward gathered at each test step over never seen environments.
This approach also starts providing a significant increase early on (after training on 50 environments).  This indicates that this approach generalizes better but also earlier than the other ones. 
We also highlight that this approach increases stability too, as shown by the smaller size of the standard error than the other approaches.
The random initialization technique is performing worse than the basic initialization. The best agent initialization performs similarly than the basic initialization.
\paragraph{Number of agents in the pool}
The number of agents in the pool (Figure \ref{init_results})  also reflects the capacity of each agent to cover a wider number of environments on which the eco-system has been trained. In Figure~\ref{init_results}, we can see that the forked agent approach is showing a bit higher number of agents in the pool compared to the best agent and the basic initialization approaches. This increase of the number of agents comes from the fact that each agent trained using the forked agent initialization embedded a better generalization to other environments. 
This means that each agent created this way covers a wider number of environments than with the other initialization techniques. The optimization technique only removes agents from the pool when all environments of an agent can be solved by another agent. The better generalization capabilities in this case makes it more difficult for one agent to fully match all the environments of another agent, leading to some overlap and an increase in the number of agents in the pool.
The random initialization technique is using an higher number of agents, but this is only due to the lack of performance of the agents initialized this way.
The basic initialization and the best agent initialization performs again similarly.
\paragraph{Number of training steps}
The forked agent approach (Figure \ref{init_results}) is clearly superior when looking at the number of training steps needed to complete the training on 500 environments, being nearly half of what is needed by the best agent and random approaches.
This is easily explained because each agent created embeds far better generalization capabilities as it has been trained on a lot more environments (as all coming from the Main Agent) before being trained on its dedicated environment. Then the additional training needed to solve its dedicated environment is far less than for the other options.
We can also see that any initialization techniques performs far better than the standard initialization originally proposed with the eco-system setup, around 3 time better for the random and best agent approach and 6 times better for the forked agent. This can be easily explained by the fact that any new agent is initialized with a Neural Network which has already been trained on similar but slightly different environment that the one on which it is trained, then a part of the learning has already been done and is transferred to the new agent.

Overall, we can see that the forked agent approach offers a big increase in term of performance as well as in term of stability when comparing the deviation error.

\section{Discussion / conclusion}
\label{discussion_conclusion}
We have explored different ways of initializing the new agents in the eco-system setup and the associated increases of performance.
The forked agent initialization approach improves significantly the generalization capabilities of the solution as well as reduces drastically the number of training cycles needed compared to the original eco-system approach.
By using agents which have been trained on multiple environments before joining the pool of policies, we have increased the generalization of each agent used in this new setting. This approach leverages the best of the two worlds, the increased generalizability of an agent which has been trained on multiple environments as well as the stability of the eco-system setting where catastrophic forgetting has been fully eliminated.
In order to enhance the performance of both the eco-system as well as the newly proposed forked agent approach, it would be interesting to see if we can find a way to predict which agents are more prone to generalize and focus on these and quickly remove the other ones from the pool. Also, trying to reduce the number of inferences by predicting if a given agent will offer good performance on a given environment without running it would give a significant performance increase to these approaches.

\end{document}